\documentclass[letterpaper, 10 pt, conference]{ieeeconf}  

\IEEEoverridecommandlockouts
\usepackage{cite}
\usepackage{amsmath,amssymb,amsfonts}
\usepackage{algorithmicx}
\usepackage{algorithm}
\usepackage{algpseudocode}
\algrenewcommand\algorithmicrequire{\textbf{Input:}}
\algrenewcommand\algorithmicensure{\textbf{Output:}}
\usepackage{graphicx}
\usepackage{textcomp}
\usepackage{xcolor}
\usepackage{pgfplots}

\usepackage{csquotes}
\usepackage{bbm}
\usepackage{tikz}
\usepackage{pgfplots}
\usepackage{xcolor}
\usepackage{subcaption}
\usepackage{multirow}
\pgfplotsset{compat=1.18}
\usetikzlibrary{positioning}
\usetikzlibrary{decorations.pathreplacing}
\usetikzlibrary{calc}
\usetikzlibrary{positioning,shapes,shadows,arrows}
\usepackage[font=small]{caption}

\pgfplotsset{
every axis/.append style={
  axis line style={->}, 
  legend style={font=\scriptsize},
  label style={font=\scriptsize},
  title style={font=\scriptsize},
  tick label style={font=\scriptsize},
  }
}

\makeatletter
\let\NAT@parse\undefined
\makeatother
\usepackage{hyperref}

\title{\LARGE \bf Interactive Robot-Environment Self-Calibration\\via Compliant Exploratory Actions}
    
\author{%
    Podshara Chanrungmaneekul$^{1^*}$, 
    Kejia Ren$^{1^*}$,
    Joshua T. Grace$^2$,
    Aaron M. Dollar$^2$,
    Kaiyu Hang$^1$
    \thanks{$^*$Equal contribution.}
    \thanks{$^1$Department of Computer Science, Rice University, Houston, TX 77005, USA. $^2$Department of Mechanical Engineering and Material Science, Yale University, New Haven, CT 06511, USA. This work was supported by the US National Science Foundation grant FRR-2133110 and FRR-2132823.}
}
\begin{document}
\maketitle
\begin{abstract}
Calibrating robots into their workspaces is crucial for manipulation tasks. Existing calibration techniques often rely on sensors external to the robot (cameras, laser scanners, etc.) or specialized tools. 
This reliance complicates the calibration process and increases the costs and time requirements.
Furthermore, the associated setup and measurement procedures require significant human intervention, which makes them more challenging to operate.
Using the built-in force-torque sensors, which are nowadays a default component in collaborative robots, this work proposes a self-calibration framework where robot-environmental spatial relations are automatically estimated through compliant exploratory actions by the robot itself. The self-calibration approach converges, verifies its own accuracy, and terminates upon completion, autonomously purely through interactive exploration of the environment's geometries. 
Extensive experiments validate the effectiveness of our self-calibration approach in accurately establishing the robot-environment spatial relationships without the need for additional sensing equipment or any human intervention.
\end{abstract}

\section{Introduction}

Robots are becoming increasingly prevalent in various industrial and household applications, undertaking tasks ranging from pick-and-place to tool manipulation within defined environments. 
Often, such robot applications require an accurate pose of the robot frame relative to a frame of the workspace to be provided by a calibration procedure before task execution. Robot calibration involves identifying the true geometrical parameters in the robot's kinematic structure and enhances absolute pose accuracy through software adjustments, avoiding alterations to the robot's mechanical structure \cite{roth1987overview}. 
While certain calibration types, such as hand-eye calibration \cite{qiu1995practical, meng2007autonomous, du2013online} or object localization \cite{villalonga2021tactile, bauza2023tac2pose}, have been previously studied, research on automatic robot-environment calibration for manipulators, a critical aspect of overall robot performance, has been relatively limited.

Traditional robot-environment calibration approaches rely on external measurement devices, which are time-consuming and require significant human intervention in the loop. 
To automate and streamline this process, various calibration methods have explored sensor modalities like vision, 1D laser sensors, touch-based techniques, tactile sensors, and probes\cite{meng2007autonomous, s20164354, petrovskaya2011global,sipos2022simultaneous, stepanova2022automatic,du2013online, saund2017touch}, complicating the calibration and limiting accessibility for users without these resources. 
When robot physical interactions are needed for calibration, such as probing points in the environment, manual operations are still often required since existing simulation-based interaction and outcome prediction \cite{saund2017touch, 9709520, 6630562} have not shown to generalize to arbitrary robot-environment setups.
\begin{figure}[!t]   
    \centerline{\includegraphics[width=0.9\columnwidth]{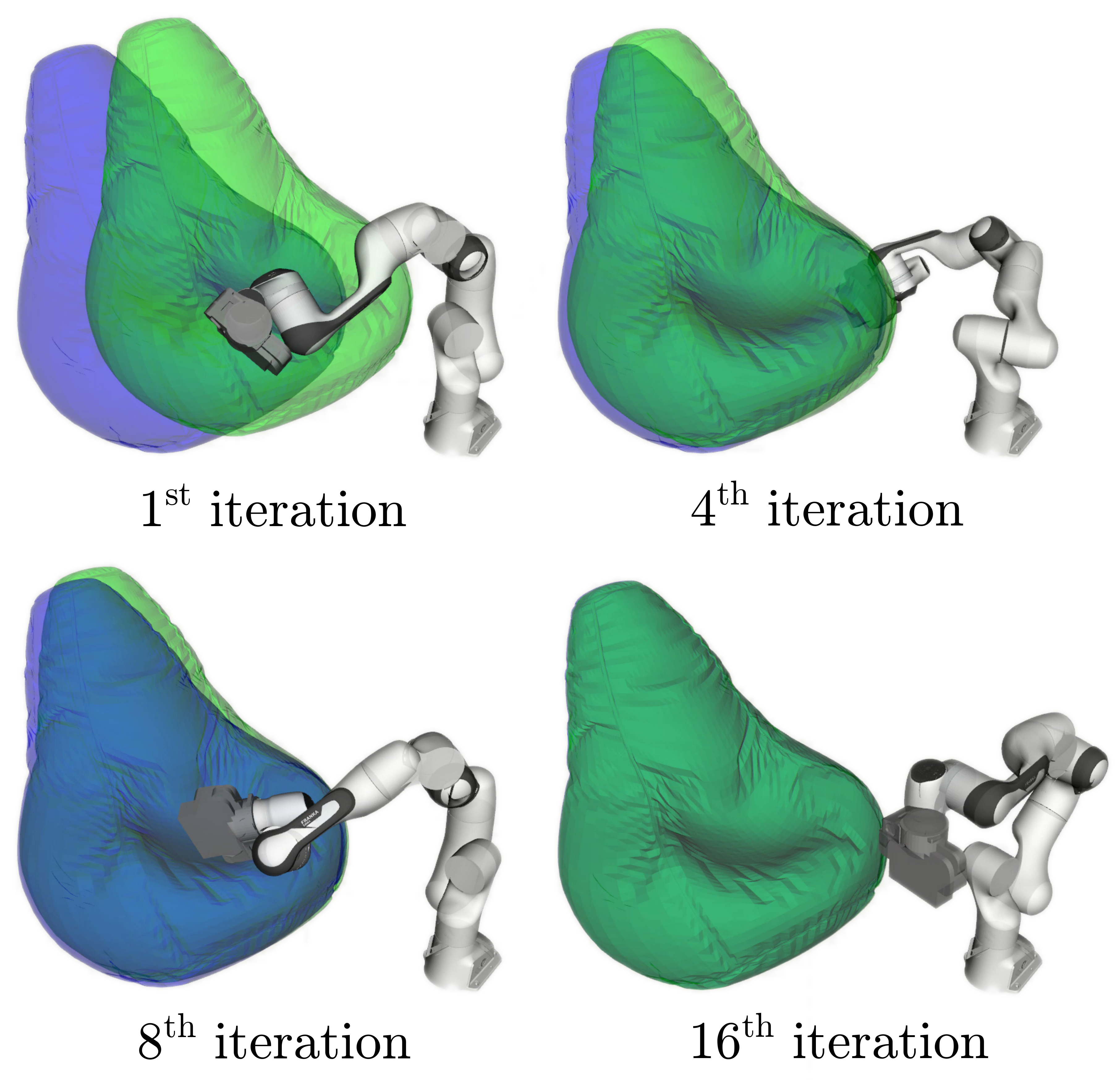}}
    \caption{Our self-calibration framework estimates the robot-environment spatial relationship via compliant exploratory actions. Visualized in green is the environment's pose as currently estimated by the robot, and in blue is the ground truth.}
\label{pic:overview}
\vspace{-0.6cm}
\end{figure}

To this end, we propose a framework for robot-environment \emph{self-calibration}, defined as a fully autonomous process for identifying the robot-environment spatial parameters only using the robot's exploratory actions without any human interventions. 
We work with setups, as exemplified in Fig. \ref{pic:overview}, without external sensors or specialized end-effectors, e.g., point contact-based probes. 
The robot-environment spatial relationship is modeled using a Particle Filter with a distribution of hypothesized poses.
Our framework uses the robot's end-effector to actively interact with the environment through touching and sliding actions to gather informative observations of contacts, refining an inaccurate initial belief iteratively and finally converging the hypotheses towards the true value.
The exploratory actions are strategically optimized to collect information that maximally contributes to the self-calibration. 
Experiments conducted in both simulation and the real world confirm the effectiveness and accuracy of our self-calibration framework. 
The results show that the proposed framework is able to precisely estimate the robot's pose relative to the environment, operating in a fully autonomous manner.

\section{Related work}
\label{related work}
\textit{Traditional calibration}: 
Traditionally, robot-environment calibration involves sensing the actual pose of the robot's end-effector and comparing this information with the poses predicted by the robot-environment kinematic models to acquire calibration data.
This process falls into two primary categories: robot calibration utilizing external measurement devices \cite{driels1993full,to2012improved, wang2012screw} and robot calibration through the imposition of physical constraints \cite{qiu1995practical, gong2000self, meng2007autonomous, du2013online}. 
External measurement-based robot calibration methods are time-consuming, challenging to operate, and rely on human experts to operate.
Many of such calibration approaches \cite{meng2007autonomous, du2013online} possess particular complexities, including the prerequisite of performing hand-eye calibration and robot exterior calibration in advance, with the added requirement for an external chessboard. 
In this work, we propose a robot-environment self-calibration process to eliminate the need for human intervention, external tools, and complex prerequisites.

\textit{Autonomous calibration}:
Previous works in autonomous calibration have explored multiple sensor modalities, including visions for end-effector tracking \cite{meng2007autonomous, du2013online, stepanova2022automatic}, 1D laser sensors to gauge distances to the environment  \cite{s20164354}, and the integration of touch-based techniques using a probe \cite{saund2017touch, petrovskaya2011global} or a poker \cite{sipos2022simultaneous} to establish point contacts. 
Tactile sensors have been employed to model contact dynamics \cite{koval2017manifold, koval2016pre, javdani2013efficient, koval2015pose, 6630562}, and tactile images have been leveraged for richer contact information \cite{villalonga2021tactile, bauza2023tac2pose}. 
A common feature among these approaches is the reliance on specific additional sensors or probes to facilitate calibration. In contrast, our work proposes an approach that allows the robot to self-calibrate \emph{as is} without extra sensors or tools. 

\textit{Next best action}:
During an autonomous calibration process, the measurement process iteratively refines the robot's calibration using dynamic interactions between the robot and the environment.
Various approaches have been employed to optimize this interaction. 
While some methods choose actions for calibration exploration by incorporating information gain \cite{saund2017touch, 9709520, 6630562}, others use entropy functions based on particle rejection principles to enhance the decision-making process during calibration \cite{saund2017touch}. 
However, these approaches select the next actions by calculating the potential information gain across all hypothesis states, thus limiting the number of samples for the robot's state representation. 
As an effective alternative, we develop a heuristic algorithm to select the most promising exploratory actions based on an analysis of the environment.
\section{Problem formulation}
\label{problem}
This work addresses a self-calibration problem for estimating the pose of a robot manipulator in an uncalibrated environment without the requirements of any external sensors such as cameras.
Specifically, we denote the fixed world frame of the static environment as $X^s \in SE(3)$, the base frame of the robot as $X^b \in SE(3)$, and the body frame of the robot as $X^e \in SE(3)$ rigidly attached on the robot's end-effector.
Given the geometric model of the static environment expressed in the fixed world frame, the estimation goal is to obtain the pose of the robot, $X \in SE(3)$, relative to the environment (i.e., the transformation from $X^s$ to $X^b$).

We approach the self-calibration stated above through Particle Filter-based probabilistic estimation.
It requires the robot manipulator to actively interact with the environment by touching and sliding to obtain informative observations of contacts between the robot and the environment.
By updating the particle beliefs based on the information obtained from observations and sequential importance resampling (SIR), the Particle Filter can approximate the distribution of the system's hidden state (i.e., the robot's pose $X$) and iteratively refine it based on new observations.
\subsection{Particle Filter-based Self-Calibration}
At each time step $t$ of the iterative process of Particle Filtering,
we denote the set of $M$ importance samples as $\mathcal{X}_t = \{X_t^1,\cdots, X_t^m, \cdots, X_t^M\}$, called particles for simplicity.
The value of each particle is a hypothesized pose of the robot, i.e., $X_t^m \in SE(3)$.
Based on the observational data and the actions taken by the robot,
the likelihood of having a given particle $X_t^m \in \mathcal{X}_t$ is given by the posterior:
\begin{equation}
    X^m_t \sim  p(X_t \mid Z_{1:t}, U_{1:t})
\end{equation}
where $X_t \in SE(3)$ is the robot's pose to be estimated at time $t$; $Z_{1:t}$ and $U_{1:t}$ are the observations and robot controls until time $t$ respectively. 

However, explicitly deriving the true posterior distribution above in real-world scenarios is often impractical.
As such, Particle Filters rely on a proposal distribution $\pi$ to iteratively approximate the real distribution: 
\begin{equation}
    \pi(X_t) \sim p(X_t \mid X_{t-1}, U_{t-1}) \pi (X_{t-1}) 
\end{equation}
With a set of particles $\mathcal{X}_t$ sampled from $\pi$, the estimated value of the robot's pose, given by the expected value of $X_t$, can be approximated through importance sampling:
\begin{equation}
\label{eq:expect}
        \mathbb{E}[X_t] = 
        \int p(X_t) X_t dX_t \approx \sum_{m=1}^M X_t^m w_t^m
\end{equation}
where $w_t^m \propto p(X_t^m) / \pi(X_t^m)$ is the normalized weight associated with each particle.

Assuming the system is Markovian, the weight $w_t^m$ can be recursively computed using Bayes' Theorem:
\begin{equation}
        w_t^m = \frac{p(Z_t \mid X_t^m) p(X_t^m \mid X_{t-1}^m, U_t)}{\pi(X_t^m \mid X_{t-1}^m, U_t, Z_t)}
\end{equation}
where $p(Z_t \mid X_t^m)$ is the observation model, and $p(X_t^m \mid X_{t-1}^m, U_t)$ is the motion model, which describes how the system transitions under the control input $U_t$.
\subsection{Motion Model and Observation Representation}
\label{sec:motion_obv}
In this work, we use the robot's touching and sliding actions to interact with the uncalibrated environment.
Since the environment is static and the robot manipulator is fixed in the environment, the robot's sliding action $U_t$ will have no impact on the robot's pose relative to the environment.
As such, we model the motion model with a multivariate Gaussian distribution independent of the robot action $U_t$:
\begin{equation}
\vspace{-0.2cm}
\label{eq:motion_model}
    p(X_t^m \mid X_{t-1}^m, U_t) \sim \mathcal{N}(X_{t-1}^m , \sigma_t \mathbb{I}) 
\end{equation}
where $\sigma_t$ scales the covariance of the Gaussian to simulate
system uncertainties.
This motion model is equivalent to perturbing each particle in the previous time step with random Gaussian noise. 

After the robot executes the action $U_t$ at time $t$, 
it will observe $Z_t$ with the readings of its internal joint encoders and force-torque (FT) sensor.
We represent this observation as a set of $J_t$ discrete contact events while the robot is executing the sliding action $U_t$, that is, $Z_t = \{(q^1_t, c^1_t), \cdots, (q_t^j, c_t^j), \cdots, (q^{J_t}_t, c^{J_t}_t)\}$.
Each contact event $(q_t^j, c_t^j) \in Z_t$ consists of a robot configuration (e.g., joint angles) $q_t^j \in \mathbb{R}^N$ where $N$ is the number of DoFs of the robot, and a binary value $c_t^j \in \{0, 1\}$ indicating whether the robot is in collision with the environment.

\begin{algorithm}[t]
    \caption{Self-Calibration using Particle Filter}
    \label{alg:self-calibration}
    \small
    \begin{algorithmic}[1]
    \Require Initial guess of the robot's pose $X_0 \in SE(3)$.
    \Ensure Estimated pose of the robot $X_t \in SE(3)$.
        \State $t \gets 0$
        \While {\textbf{not} terminated} \hfill \Comment{Sec. \ref{verification:verification}}
            \State $ t \gets t+1$
            \State $\mathcal{X}_t \gets \emptyset$ \hfill \Comment{Set of Particles Initialized as Empty}
            \State $U_t \gets \Call{ActionSelection}{U_{1:t-1}}$ \hfill \Comment{Sec. \ref{action:optimization}}
            \State $Z_t \gets \Call{Execute}{U_t}$ \hfill \Comment{Observation from FT Sensor}
            \For {$m = 1, \cdots, M$}
                \State $X^m_t \sim \mathcal{N}(X_{t-1}^m , \sigma_t \mathbb{I})$ \hfill \Comment{Motion Model, Eq.~\eqref{eq:motion_model}}
                \State $w_t^m \gets$ \text{Eq.~\eqref{eq:weight}} \hfill \Comment{Weight Update in Sec.~\ref{geometric:eval}}
            \EndFor
            \State $\Call{Normalize}{\left\{w_t^1,\cdots w_t^m\right\}}$
            \For {$m = 1, \cdots, M$}
                \State Sample $i$ with probability $\propto w_t^i$ \hfill \Comment{Resampling}
                \State $\mathcal{X}_t.\text{append}(X^i_t)$ \hfill \Comment{New Particles}
            \EndFor
        \EndWhile
        \State \Return $X_t \gets \sum_{m=1}^M X_t^m w_t^m$ \hfill \Comment{Eq.~\eqref{eq:expect}}
    \end{algorithmic}
\end{algorithm}
The implementation of Particle Filters for robot self-calibration is detailed in Alg.~\ref{alg:self-calibration}.
Given an initial guess of the robot's pose $X_0 \in SE(3)$, the robot needs to actively determine the sliding action $U_t$ that it will perform to collect observations, as will be detailed in Sec.~\ref{action}.
The observations will then be used to evaluate and update the weights of particles for resampling, as described in Sec.~\ref{geometric}.
By repeating this procedure, the robot will iteratively refine the approximated distribution of the robot's pose $X_t$ until the estimated robot's pose converges close to its ground truth, based on sets of particles and observations that will be introduced in Sec.~\ref{verification}.


\section{Update Rule of Particle Weights}
\label{geometric}
A critical factor for the success of Particle Filtering is the particle weight assignment based on the observations.
A robust weight evaluation can greatly facilitate the efficient resampling of particles.
As discussed in Sec.~\ref{sec:motion_obv}, we observe binary contact events detected by the internal FT sensor of the robot.
Hence, the observation model $p(Z_t \mid X_t^m)$ is the probability of contact modeled with the hypothesized distance between the robot's end-effector and the environment.

In Sec.~\ref{geometric:contact}, we implement an efficient evaluation of the hypothesized distance between the robot's end-effector and the environment by using a Signed Distance Field (SDF).
Then, based on this hypothesized distance, the rule for updating the particle weights $w_t^m$ is detailed in Sec.~\ref{geometric:eval}.

\subsection {Geometric Representations of Robot and Environment}
\label{geometric:contact}

\begin{figure}[t]\centerline{

    \includegraphics[width=0.9\columnwidth,keepaspectratio]{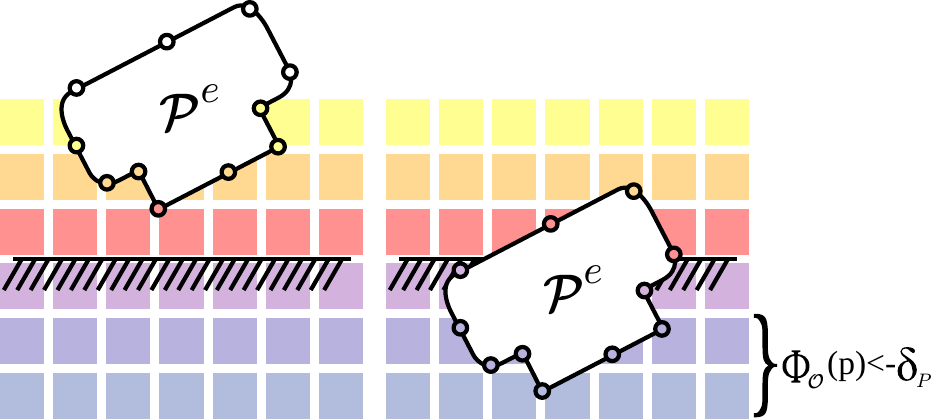}}
    \caption{The geometry of the robot's end-effector is approximated as a point cloud $\mathcal{P}^e$.
    The grid represents a voxelized cache SDF $\Phi_\mathcal{O}$ where each voxel's color corresponds to the distance to the surface boundary $\partial O$ of the environment. 
    The left figure shows a non-contact scenario where the signed distance $d_t^{m,j}$ between the end-effector and the environment surface is positive; the right figure shows a penetration scenario where the signed distance is less than a threshold, $d_t^{m,j} < -\delta_P$.}
    \label{pic:contact_rep}
    \vspace{-0.3cm}
\end{figure}

We consider the scenario where the robot manipulator operates in a static environment.
We denote the geometric model of the environment by $\mathcal{O} \subset \mathbb{R}^3$ and its surface boundary by $\partial\mathcal{O} \subset \mathcal{O}$, both expressed in the fixed world frame $X^s$.
The computation of the distance between two exact geometric models in the 3D space can be expensive, especially for complex object shapes.
In Particle Filtering, such computation needs to be performed for each particle, which limits the algorithm's speed.
To this end, the geometric model of the environment is approximated by a volumetric grid.
We represent the geometry of the robot's end-effector by a finite point cloud $\mathcal{P}^e$, containing a set of $L$ pre-sampled points on the surface of the end-effector denoted by $\mathcal{P}^e = \{p_1, \cdots, p_l, \cdots, p_L\}$.
The position of each point $p_l \in \mathbb{R}^3$ is given in the body frame of the robot $X^e$ attached to the end-effector.
Then, the distance between the end-effector and the environment can be approximated by the distance between the nearest point in $\mathcal{P}^e$ to the volumetric model of the environment.
To further speed up the distance computation, we generate a discrete Signed Distance Field (SDF)~\cite{v008a019} with a resolution of $\Delta_{sdf}$ for the environment, as illustrated in Fig.~\ref{pic:contact_rep}.
The SDF of the environment $\mathcal{O}$ is mathematically represented by a function $\Phi_\mathcal{O}: \mathbb{R}^3 \to \mathbb{R}$, which returns the signed distance to the surface boundary $\partial\mathcal{O}$ of the environment given a query point in the world frame $X^s$.

With the approximate geometric representations, distance computations can be efficiently performed.
Given a hypothesized robot's pose (i.e., a particle value) $X_t^m$ and an observed joint configuration of the robot $q_t^j \in \mathbb{R}^N$, the hypothesized distance between the robot's end-effector and the environment $d_t^{m, j}$ can be quickly obtained by querying the discrete SDF:
\begin{equation}
\label{eq:hypo_dist}
    d_t^{m, j} = \underset{p_l \in \mathcal{P}^e}{\min} \Phi_\mathcal{O} \left(X_t^m \cdot \Gamma(q_t^j) \cdot p_l\right)
\end{equation}
where $\Gamma: \mathbb{R}^N \to SE(3)$ is the forward kinematics of the robot that computes the transformation from the base frame $X^b$ to the end-effector's body frame $X^e$ of the robot given a joint configuration;
$X_t^m \cdot \Gamma(q_t^j) \cdot p_l$ is the transformation chain that transforms the coordinates of a point $p_l \in \mathcal{P}^e$ (on the surface of the end-effector) in the end-effector's body frame $X^e$ to the fixed world frame $X^s$, where $X_t^m$, $\Gamma(q_t^j)$ and $p_l$ are all given by their homogeneous representations. 

\subsection {Particle Weight Evaluation}
\label{geometric:eval}
At each time step $t$ of Particle Filters, the particle weight $w_t^m$ is updated based on the received observation $Z_t$ consisting of $J_t$ pairs of joint configuration and detected contact $(q_t^j, c_t^j)$.
When a contact is detected, i.e., $c_t^j = 1$, we use a Gaussian with a preset variance $\sigma^2_P$ to model the likelihood of observing this contact.
Specifically, if the hypothesized distance between the end-effector and the environment $d_t^{m, j}$ for particle $X_t^m$ is close to zero based on the calculation in Eq.~\eqref{eq:hypo_dist}, the likelihood of observing a contact for this particle is large, and we assign a large weight to this particle.
In contrast, if $d_t^{m, j}$ is large, which conflicts with the observed contact, we will assign a small value to the particle weight.

When no contact is observed, $c_t^j = 0$, we introduce a rejection mechanism for particles significantly inconsistent with the observation.
If the hypothesized distance $d_t^{m, j}$ associated with a particle is less than a penetration distance threshold $-\delta_P$, as shown in Fig.~\ref{pic:contact_rep} (right), it indicates the robot is predicted to greatly penetrate the environment.
Then, the weight of this particle will be multiplied by an arbitrarily small value $\epsilon$.
The weight of each particle $w_t^m$ is then aggregated as the product of all observed contact events in $Z_t$ stemming from a sliding action $U_t$.
The updated rule for particle weights assignment under different cases discussed above is given by:
\begin{equation}
\label{eq:weight}
    w_t^m = \prod_{j=1}^{J_t}
    \begin{cases}
    \frac{1}{\sqrt{2 \pi \sigma_P^2}} \exp{-\frac{|d_t^{m.j}| ^2}{2 \sigma_P}} & c_t^j = 1\\
    \epsilon &  c_t^j = 0 \wedge d_t ^{m,j} < - \delta_P \\
    1 &  \text{otherwise}
    \end{cases}
\end{equation}
After the particle weights are updated, they are normalized by $w_t^m = w_t^m / \sum_{m=1}^M w_t^m$ (Line 11 of Alg.~\ref{alg:self-calibration}).
As such, we can use the weights as the sampling probabilities to resample particles.


\section{Robot Compliant Action Selection}
\label{action}

Each robot action entails a time cost; hence, it is imperative to reduce the number of actions needed for accurate self-calibration and to ensure that every action captures a maximal amount of information.
To this end, we use sliding actions of the robot to gather as much observation information as possible, as described in Sec.~\ref{action:action}.
In Sec.~\ref{action:optimization}, we develop a strategy to select the most promising sliding action from multiple candidates heuristically.

\subsection{Compliant Sliding Action}
\label{action:action}



The robot will inevitably have unexpected collisions with the environment while interacting with it 
through sliding actions due to the inaccuracy of the self-calibration 
and real-world uncertainties, such as control errors.
In response to such uncertainties, enabling robot compliance becomes imperative to facilitate safe interactions and maintain continuous contact with the uncalibrated environment.


Specifically, a sliding action $U_t$ is associated with a reference contact $ (r_t, \hat{n}_t)$ on the environment surface, which includes a contact location $r_t \in \mathbb{R}^3$ and its corresponding normal vector $\hat{n}_t \in \mathbb{R}^3$, both expressed in the fixed world frame $X^s$.
To execute the sliding action $U_t$, the robot moves close to and above the reference location $r_t$. 
Following this, to ensure safe interactions, the robot utilizes Cartesian impedance control to guide its end-effector to gently touch the environment surface and detect contact.

\begin{figure}[t]
    \centerline{\includegraphics[width=0.95\columnwidth,keepaspectratio]{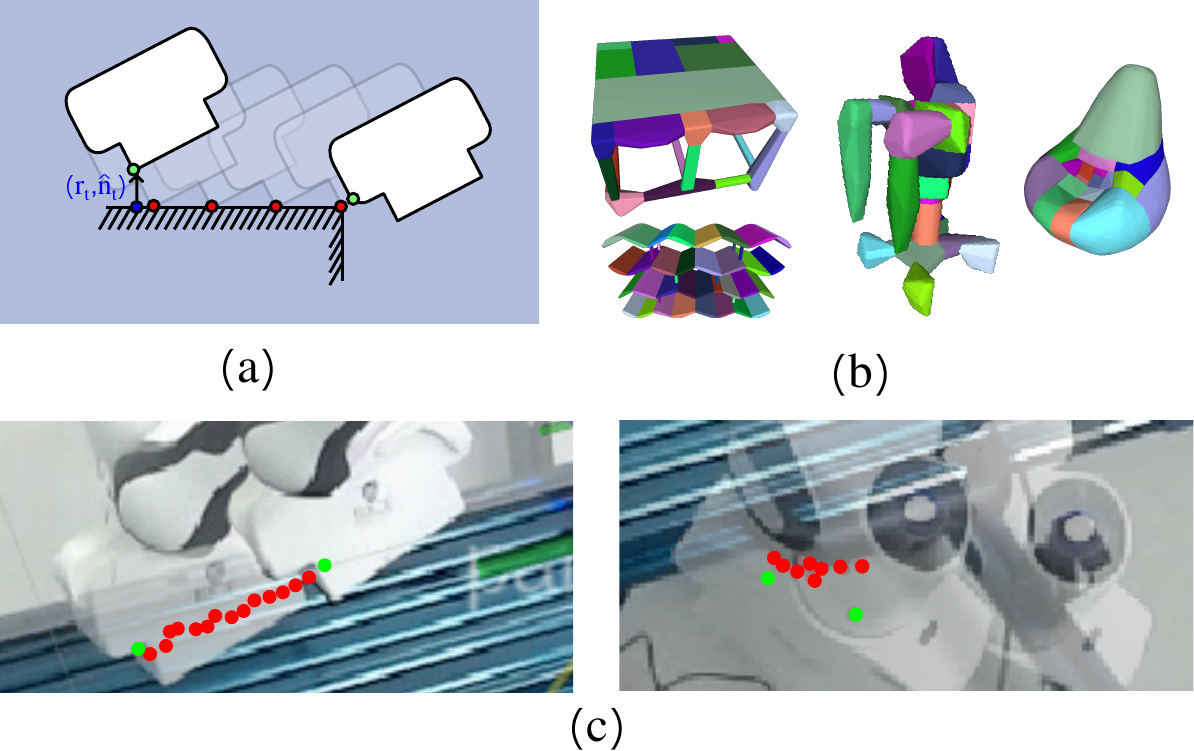}}
    \caption{(a) An illustrative plot of a sliding action $U_t$.
    The blue dot is the reference contact location $r_t$ on the environment surface, and the black arrow represents the normal vector $\hat{n}_t$ at the contact location.
    The recorded observations for contact and non-contact are shown by the red and green dots, respectively. (b) Examples of convex segmentation for objects in Fig.~\ref{pic:objects}. (c) Examples of a real robot executing two different sliding actions.
    }
    \label{pic:sliding}
    \vspace{-0.3cm}
\end{figure}

Upon contact detection by the FT sensors of the robot, the robot transitions into a sliding motion along the surface, maintaining contact in a direction perpendicular to the surface normal vector $\hat{n}_t$, as shown in Fig.~\ref{pic:sliding} (a). 
The sliding continues until the robot loses contact with the environment or exceeds a pre-defined maximum sliding distance. 
Throughout the sliding action, the robot collects contact events $(q_t^j, c_t^j) \in Z_t$ at fixed distance intervals, 
regardless of the contact status.
Fig.~\ref{pic:sliding} (c) showcases two different sliding actions executed by a real robot.

\subsection{Action Optimization}
\label{action:optimization}

To further maximize the information gained from a single robot action,
we propose a strategic approach to optimize the action selection to enhance estimation efficiency.
The objects in the environment are divided into convex segments using CoACD \cite{wei2022approximate}, as shown in Fig.~\ref{pic:sliding} (b).
Each segment captures the local geometric information of the environment and serves as a focal point guiding the estimation process to concentrate on different parts of the environment.
For each segment, we pre-sample a set of $D$ possible contact candidates $\mathcal{D} = \{(r_1, \hat{n}_1), \cdots, (r_d, \hat{n}_d), \cdots, (r_D, \hat{n}_D) \}$. 
Each contact candidate $(r_d, \hat{n}_d)$ consists of a contact location $r_d \in \mathbb{R}^3$ on the surface of the segment and a normal vector $\hat{n}_d \in \hat{\mathbb{R}^3}$ at the contact location.
As discussed in Sec.~\ref{action:action}, each robot sliding action $U_t$ is associated with a reference contact $(r_t, \hat{n}_t)$, selected from the pre-sampled contact candidates $\mathcal{D}$ of all segments.


When the robot selects a sliding action $U_t$ at time $t$ to optimize the information gathering, it will retrieve all the previous actions $U_{1:t-1}$ that the robot has executed.
The robot will strategically select an action that explores different geometric features of the environment unsampled in the previous steps.
Specifically, the robot first randomly picks a segment of the environment.
For each contact candidate $(r_d, \hat{n}_d) \in \mathcal{D}$ in this segment, we define a local sparsity $\rho_\mathcal{D}(d)$, quantified as the angular difference between its normal vector $\hat{n}_d$ and the normal vectors of the reference contacts associated with all previous actions, i.e., $\hat{n}_{1:t-1}$:
\begin{equation}
    \rho_\mathcal{D}(d) = \underset{\tau = 1, \cdots,  t-1}{\min} \cos^{-1} \hat{n}_d \cdot \hat{n}_\tau
\end{equation}

The strategy is to select the contact candidate with the maximum local sparsity
as the reference contact for $U_t$, 
i.e., $(r_t, \hat{n}_t) = (r_{d^*}, \hat{n}_{d^*}) \in \mathcal{D}$ where $d^*$ is calculated by
\begin{equation}
    d^* = \underset{d = 1, \cdots, D}{\arg\max} \text{ } \rho_\mathcal{D} (d)
\end{equation}

The robot then moves its end-effector to the selected location $r_{d^*}$, uses Cartesian impedance control to touch the environment surface, and slides its end-effector in a direction perpendicular to $\hat{n}_{d^*}$ to execute $U_t$, as described in Sec.~\ref{action:action}.
In cases where the selected contact candidate results in an infeasible action due to the reachability of the robot, it will attempt the next best candidate until a feasible one is found. 
Alternatively, there is a small probability of switching to another segment to explore alternate contact options and select a reference contact from the candidates on the new segment. 
This adaptive strategy ensures the viability of the selected contact and the sliding action associated with this contact.

\section{Convergence Criteria and Verification of the Estimation}
\label{verification}
We assume the ground truth of the robot's pose is not obtainable, and no other external sensors are available to verify the quality of the estimation during the self-calibration. Thus, we need an automatic mechanism to self-measure the estimation convergence and end the self-calibration when needed.
In Sec.~\ref{verification:verification}, we define criteria to measure the confidence of the estimation based on the statistical state of the particles and implement an adaptive strategy to control the spreading of the particles under different conditions to avoid overconfidence in a potentially false estimation. 


\subsection{Confidence and Convergence Criteria}
\label{verification:verification}

We define three binary criteria to measure the confidence of the estimation, based on the set of particles $\mathcal{X}_t$ at the current step or all particles in the previous steps $\mathcal{X}_{1:t}$:

    1) \emph{Particle Confidence:}
    $\mathcal{C}(\mathcal{X}_t) \in \{0, 1\}$ is determined by examining the variance of the particles' values: 
    \begin{equation}
    \label{eq:particle_conf}
        \mathcal{C}(\mathcal{X}_t) = 
        \begin{cases}
            1 & \sigma(\mathcal{X}_t) < \theta_V\\
            0 & \text{otherwise}
        \end{cases}
    \end{equation}
    where the dimension-wise variance $\sigma(\mathcal{X}_t) \in \mathbb{R}^6$ independently evaluates the variance of all six dimensions of $\mathcal{X}_t$, after converting the representation of each $X_t^m \in SE(3)$ in the set $\mathcal{X}_t$ to a six-dimensional vector (i.e., x, y, z, roll, pitch, and yaw).
    When the variance is below a threshold $\theta_V \in \mathbb{R}^6$, indicating all particles tend to converge to the same estimate, the Particle Filter estimator is confident about the estimation.

    2) \textit{Particle Stability:}
    $\mathcal{S}(\mathcal{X}_{1:t}) \in \{0, 1\}$ is evaluated based on the particle values of all previous steps $\mathcal{X}_{1:t}$:
    \begin{equation}
    \label{eq:particle_stab}
    \begin{aligned}
        \mathcal{S}(\mathcal{X}_{1:t}) &= 
        \begin{cases}
            1 & \mathrm{ran}_M < \epsilon_M \wedge \mathrm{ran}_V < \epsilon_V\\
            0 & \text{otherwise}
        \end{cases}
    \end{aligned}
    \end{equation}
    where $\mathrm{ran}_M \in \mathbb{R}^6$ is the dimension-wise range of $X_{t-h:t}$, the estimates (i.e., the weighted average of particle values) over the past $h$ iterations, and $\mathrm{ran}_V \in \mathbb{R}^6$ is the range of $\left\{\sigma(\mathcal{X}_{t-h}), \cdots, \sigma(\mathcal{X}_t)\right\}$, the dimension-wise variances of the particle values in the past $h$ iterations.
        
    In other words, this ensures both the weighted average and the variance of the particle values do not change much during the past $h$ steps. If their ranges $\mathrm{ran}_M$ and $\mathrm{ran}_V$ are smaller than their respective thresholds $\epsilon_V$, $\epsilon_M \in \mathbb{R}^6$, the estimation is considered consistent over the time window, meaning the particles are stable.

    3) \textit{Particle Consistency:}
    $\mathcal{V}(\mathcal{X}_t, Z_t) \in \{0, 1\}$ is used to evaluate whether the estimation is consistent with the most recent observation $Z_t$.
    Assuming the estimated robot pose determined by the weighted average of particle values $X_t$ is correct, for each observed joint configuration $q_t^j$, we calculate an estimated distance $d_t^{E, j}$ between the robot's end-effector and the environment using Eq.~\eqref{eq:hypo_dist} with $X_t$ instead of $X_t^m$.
    For all observed events in $Z_t$, if the estimated distance $d_t^{E, j}$ is not consistent with the detected contact $c_t^j$, thresholded by a distance $\delta_E$, then the Particle Consistency criterion will be considered unsatisfied.
    This criterion is met only when the estimated distance $d_t^{E, j}$ is consistent with the detected contact for all observed events $(q_t^j, c_t^j) \in Z_t$:
    \begin{equation}
    \label{eq:particle_cons}
    \begin{aligned}
        \mathcal{V}(\mathcal{X}_t, Z_t) &= 
        \begin{cases}
            0 & \exists (q_t^j, c_t^j) \in Z_t, \\
              & \text{s.t. } (c_t^j = 1 \wedge d_t^{E, j} > \delta_E) \\
              & \text{or } (c_t^j = 0 \wedge d_t^{E, j} \leq -\delta_E)\\
            1 & \text{otherwise}
        \end{cases}\\
        d_t^{E, j} &= \underset{p_l \in \mathcal{P}^e}{\min} \Phi_\mathcal{O} \left(X_t \cdot \Gamma(q_t^j) \cdot p_l\right)
    \end{aligned}
    \end{equation}

When all three criteria are met for five consecutive iterations, the Particle Filter will terminate and return the weighted average of the current particle values as the final estimation for the robot's pose.

\subsection{Adaptive Particle Spreading}
\label{verification:motion_model}

As the robot's pose is time-invariant in the motion model of the system, as introduced in Sec.~\ref{sec:motion_obv}, the motion model is a random Gaussian perturbation scaled by $\sigma_t$. 
When $\sigma_t$ is large, the particles will disperse between iterations, whereas when $\sigma_t$ is small, the particles will become more stable and more easily trapped at a local minimum.
Therefore, to better deal with scenarios where the particles distribute differently,  we adaptively adjust the scale of the variance in the motion model, i.e., the value of $\sigma_t$:

\begin{equation}
\label{eq:sigma_t}
    \sigma_t = 
    \begin{cases}
        \alpha \sigma_{t-1}  & \mathcal{C}(\mathcal{X}_t) \wedge \lnot \mathcal{V}(\mathcal{X}_t, Z_t)\\
        \beta \sigma_{t-1} & \mathcal{S}(\mathcal{X}_{1:t}) \wedge \mathcal{V}(\mathcal{X}_t, Z_t)\\
        \sigma_{t-1} & \text{otherwise}
    \end{cases}
\end{equation}
where these criteria was defined in Sec.~\ref{verification:verification}$; \alpha > 1$ and $\beta < 1$ are scaling factors.

For scenarios where the particles have confidence in their estimation, yet the estimation conflicts with the recent observation data, i.e., $\mathcal{C}(\mathcal{X}_t) \wedge \lnot \mathcal{V}(\mathcal{X}_t, Z_t)$, the particles are likely trapped at a local minimum.
To address this, we introduce larger divergence to the particle distribution by increasing $\sigma_t$, allowing for broader exploration and preventing confinement of the estimation to local minima.


On the contrary, if particles remain stable and the current estimation aligns with the observation (i.e., $\mathcal{S}(\mathcal{X}_{1:t}) \wedge \mathcal{V}(\mathcal{X}_t, Z_t)$), it indicates that $\sigma_t$ should be decreased to densify the particles, to enhance estimation confidence by reducing the size of the hypothesized distribution. 
This scaling approach ensures adaptability in scenarios where the estimation conflicts with the observations and fine-tunes the particle distribution for improved estimation accuracy and robustness.

\section{Experimental Evaluation}
\label{experiment}

\begin{figure*}[th]
    \centerline{\input{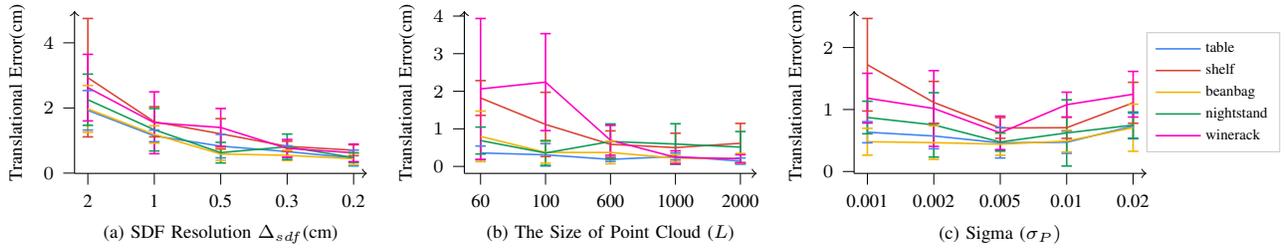}}
    \caption{Self-calibration performance is evaluated in simulation with 100,000 particles. The reported data illustrate the effects of: (a) SDF resolution, (b) end-effector point cloud size $L$, and (c) $\sigma_P$ (a particle weight evaluation parameter). 
    The results, averaged over 5 experimental runs for each environmental object, showcase the final translational error (cm), with error bars representing the standard deviation.
    \label{graph:sim_test}}
    \vspace{-0.3cm}
\end{figure*}
In this study, we thoroughly evaluate the proposed framework for robot self-calibration and investigate its performance in both simulated and real-world scenarios. 
The experiments are conducted in the Gazebo simulator \cite{koenig2004design} and on a real Franka Emika Panda robot arm. 
In the real world, the system monitors the robot joints forces and torques, which trigger collision detection when the values surpass a certain threshold. 
Throughout the experiments, the robot is equipped with a 3D-printed end-effector cover that mirrors the shape of a typical Franka gripper and serves as a protective measure to prevent damage to the robot.

\begin{figure*}[th]
    \centerline{
        \begin{minipage}[b]{.33\columnwidth}
        \includegraphics[width=\linewidth,keepaspectratio]{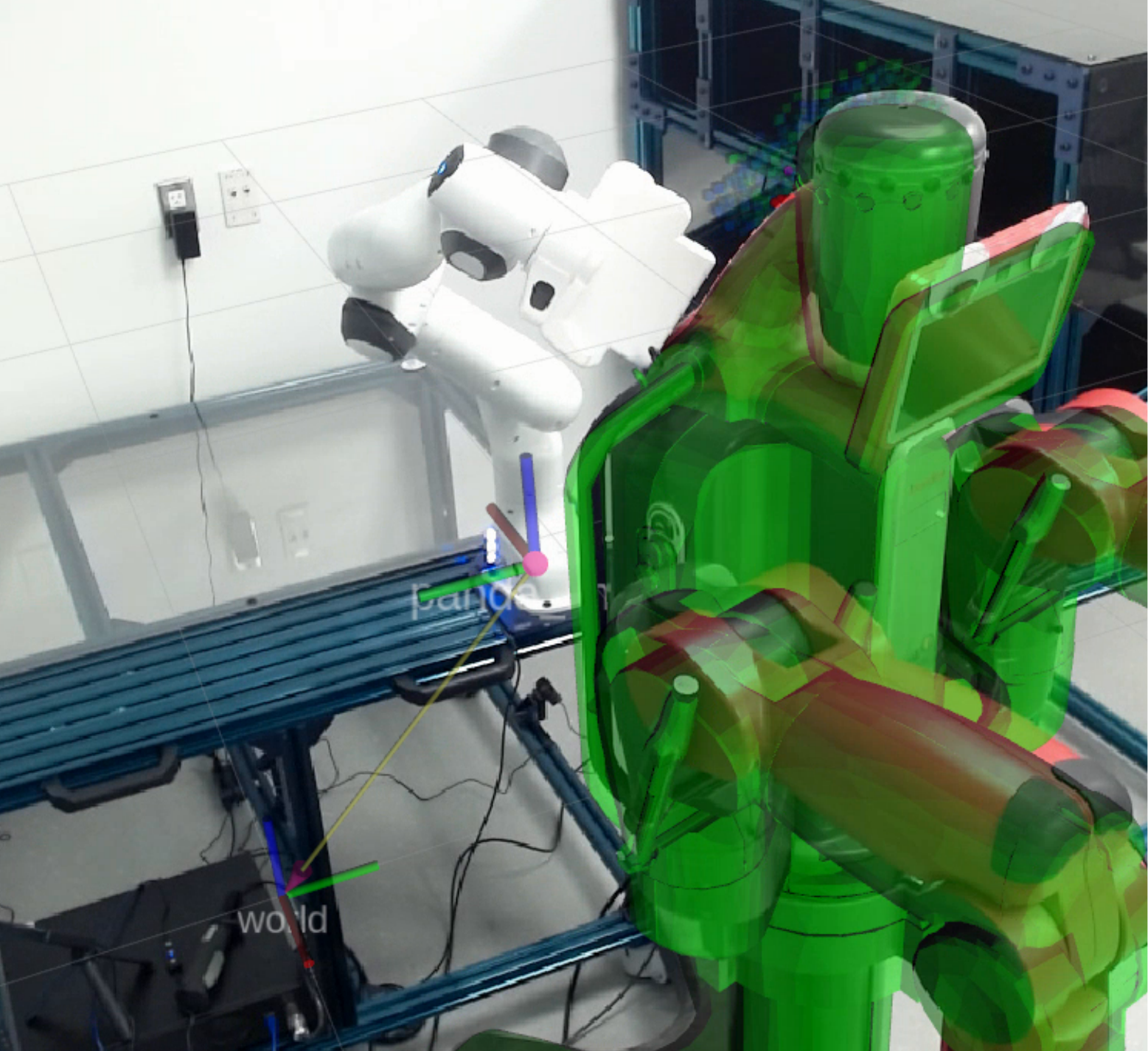}
        \end{minipage}
        \begin{minipage}[b]{.33\columnwidth}
        \includegraphics[width=\linewidth,keepaspectratio]{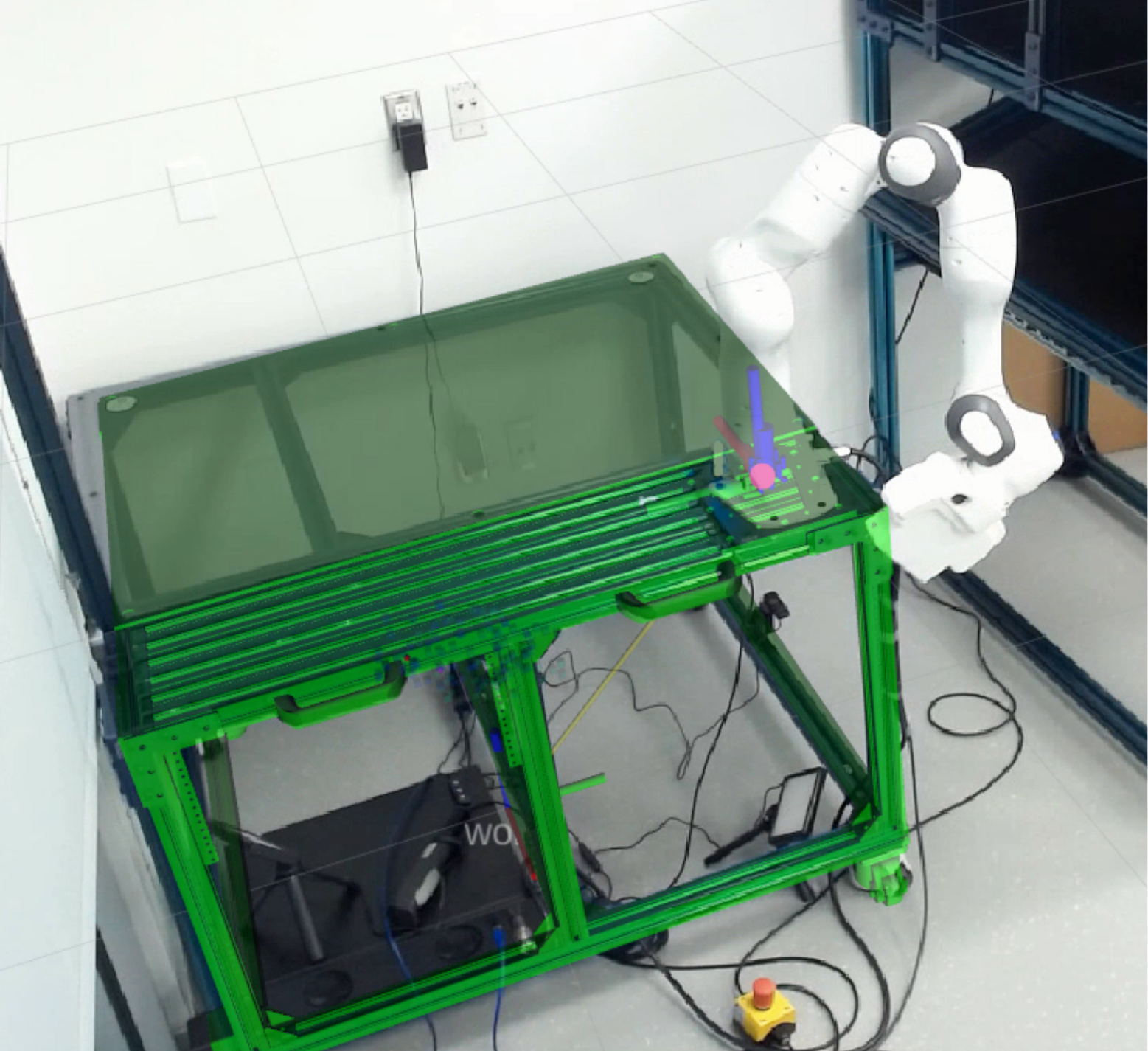}
        \end{minipage}
        \begin{minipage}[b]{.33\columnwidth}
        \includegraphics[width=\linewidth,keepaspectratio]{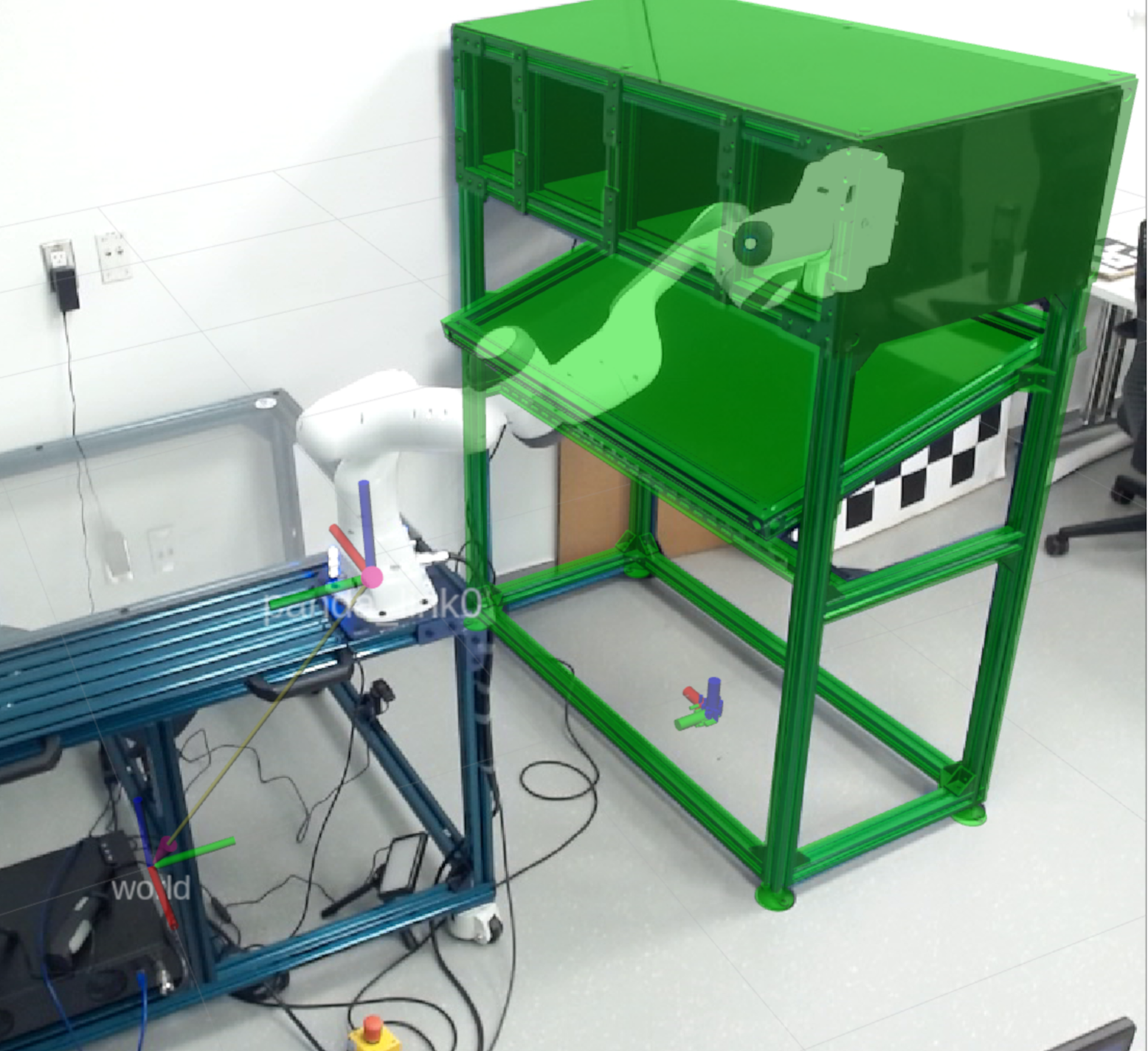}
        \end{minipage}
        \begin{minipage}[b]{.38\columnwidth}
        \includegraphics[width=\linewidth,keepaspectratio]{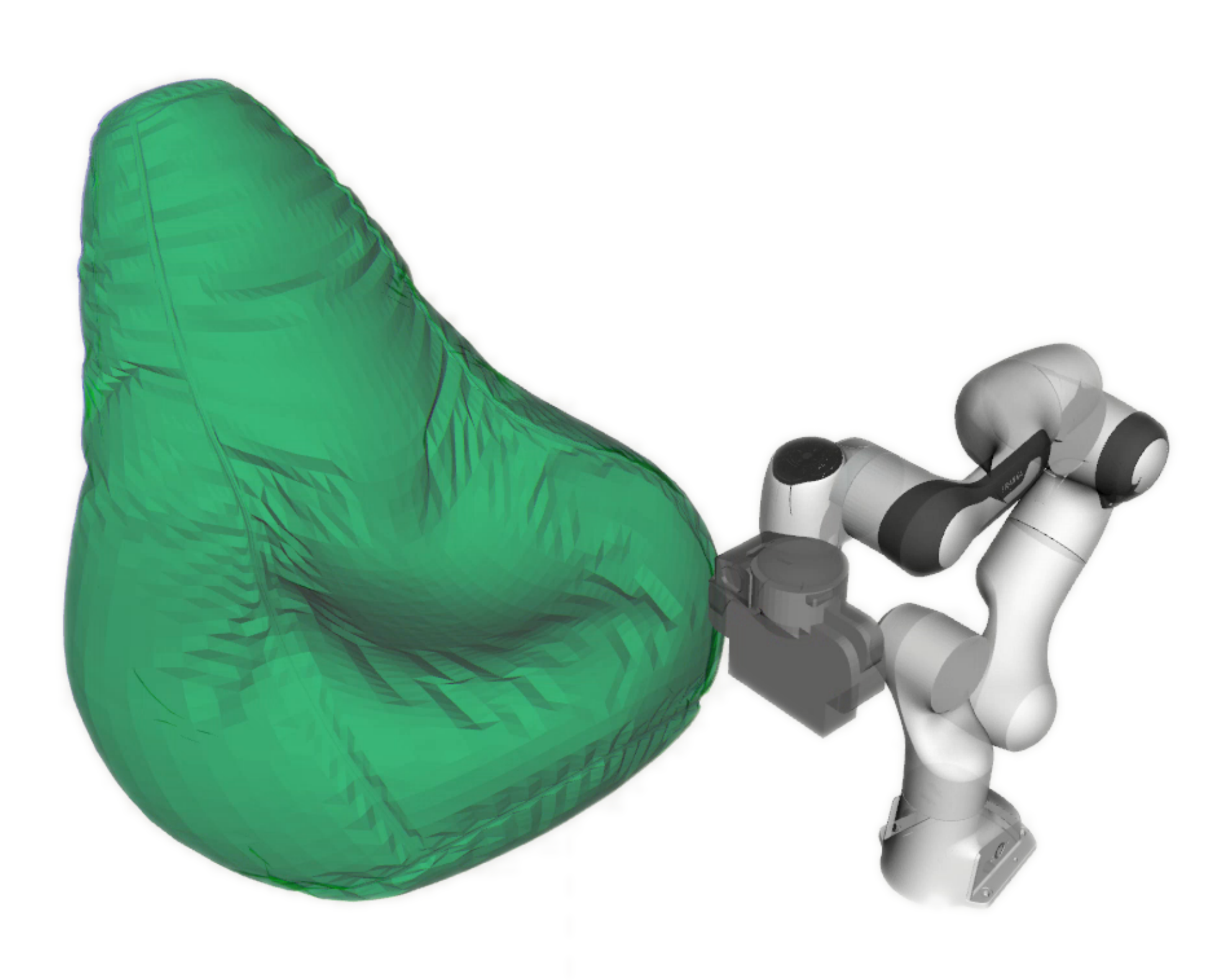}
        \end{minipage}
        \begin{minipage}[b]{.23\columnwidth}
        \includegraphics[width=\linewidth,keepaspectratio]{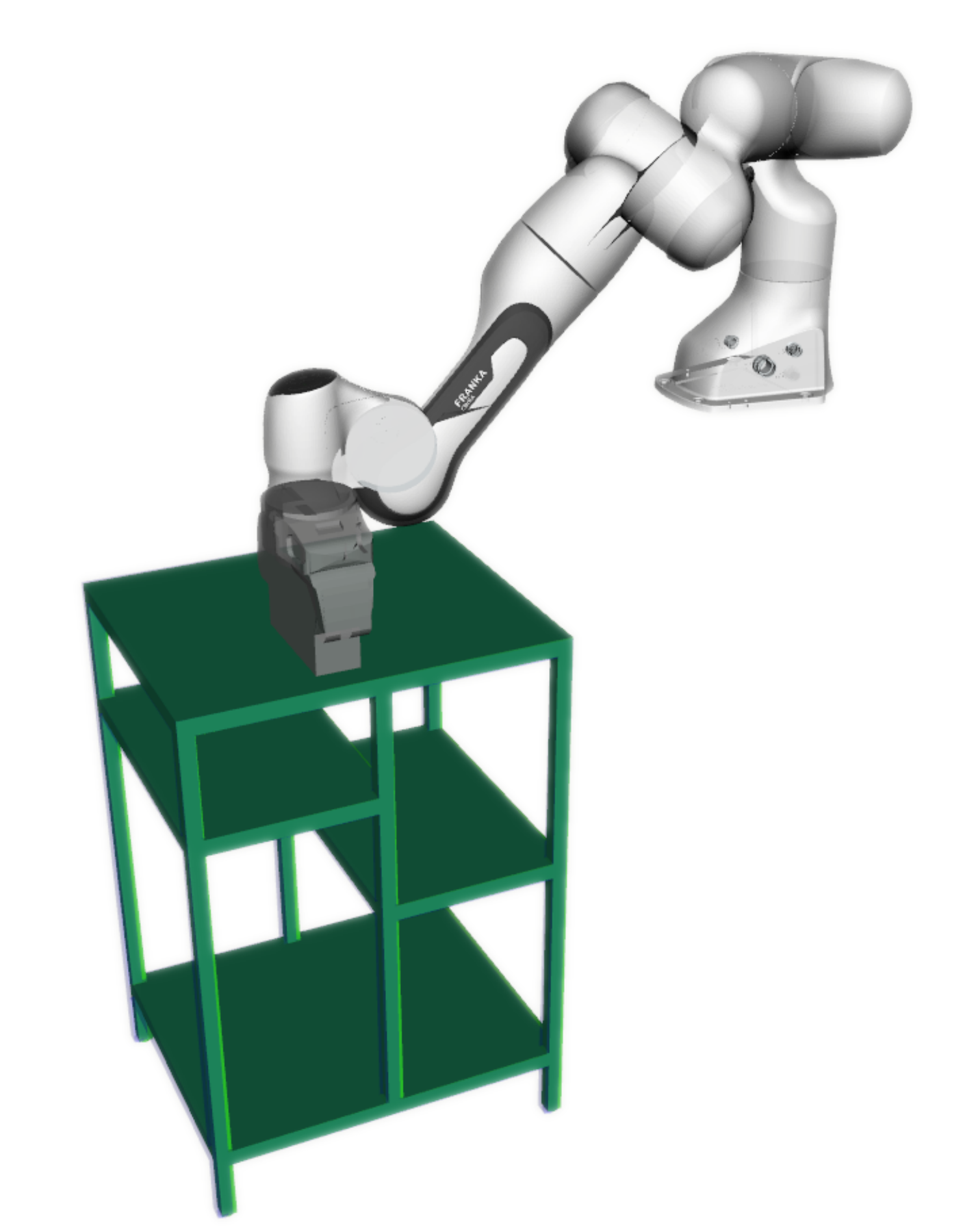}
        \end{minipage}
        \begin{minipage}[b]{.38\columnwidth}
        \includegraphics[width=\linewidth,keepaspectratio]{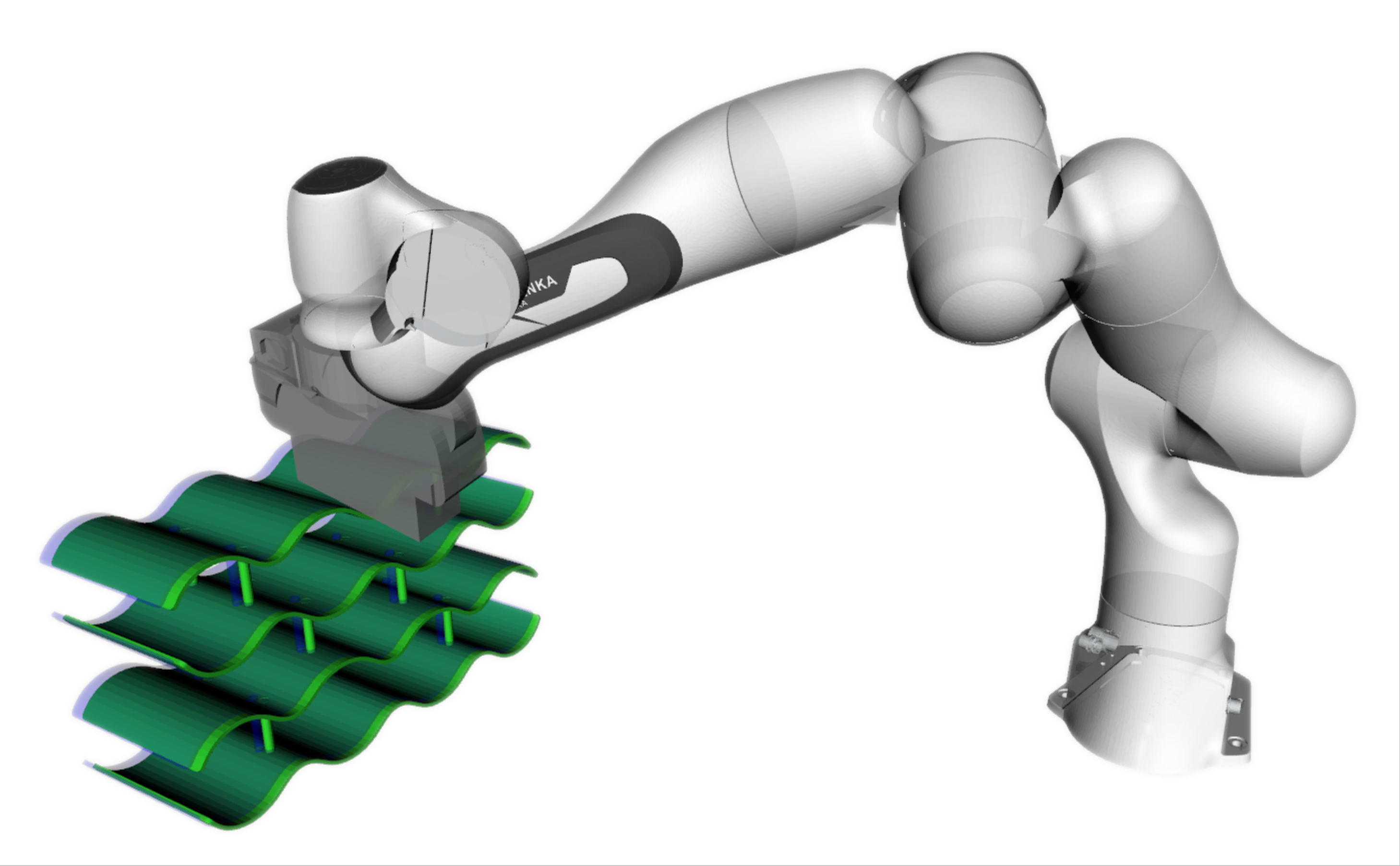}
        \end{minipage}
        }
    \caption{Environmental objects used in experiments: Baxter (real), table (sim/real), shelf (sim/real), beanbag (sim), nightstand (sim), winerack (sim).
    The green overlays are the objects' mesh models, displayed relative to the robot's base frame based on the robot's self-calibrated pose.}
    \label{pic:objects}
    \vspace{-0.3cm}
\end{figure*}

\subsection{Experiments in Simulation}
\label{experiment:sim}
As defined in Sec. \ref{problem}, we task our framework to self-calibrate the robot's pose with touching and sliding actions. 
We simulate an initial transformation error as $X - X_0 \sim \mathcal{U}(-e, e)$ where $\mathcal{U}$ is a uniform distribution and $e$ indicates the interval size of the distribution. 
The initial translational error is simulated to be within $\pm$ 15 cm, while the initial rotational error is simulated to be within $\pm$ 0.15 rad ($8.6^\circ$). 
The number of sampled particles is fixed at $M = 100,000$. 
We conduct five experiments with different initial $X_0$ values to self-calibrate the robot against each of the environmental objects, including a table, a shelf, a rigid object with the shape of a beanbag, a nightstand, and a winerack, as illustrated in Fig. \ref{pic:objects}. 
For each object, we sample a total of $1,000$ contact candidates.

There are two major features to be highlighted in the experiments:
1)  Without any tactile sensors for determining the contact location, our framework relies on accurate distance calculations between the hypothesized end-effector and the environment, which is crucial for achieving precise self-calibration.
2)  After each measurement, an efficient method is required to filter out incorrect particles based on observations.
Our framework effectively addresses these challenges, as demonstrated in the experiments detailed below. 
The self-calibration accuracy is evaluated by comparing the final translational and rotational errors, calculated as the difference between the expected value of $X_t$ from Eq. (\ref{eq:expect}) and the ground truth $X$. 
Due to the similarity between translational and rotational errors, here we focus on the experimental results of translational errors. 
Fig. \ref{graph:sim_test} presents average translational errors for each experimental setup, with error bars indicating the standard deviation.

\subsubsection{Geometric Representation Analysis}
\label{experiment:sdf}

This experiment investigates the influence of geometric representation resolution on self-calibration accuracy reflected by the final translation error. 
We show that self-calibration improves its accuracy with more accurate representations of the environment and the robot's end-effector.

We evaluate the voxel grid size of the signed distance field (i.e., $\Delta_{sdf}$ in Sec. \ref{action:optimization}) at 0.2, 0.3, 0.5, 1, and 2 cm, all with an end-effector point cloud size $L = 600$.
The voxel size affects the precision of distance calculations between the end-effector and the surrounding environment, directly affecting the particle updates during self-calibration. 
The results in Fig. \ref{graph:sim_test}(a) show a direct correlation between voxel size and calibration accuracy. 
Specifically, smaller voxel sizes consistently lead to improved calibration outcomes. 
For the voxel grid size of 0.2 cm, the average final translational error is $0.53 \pm 0.22$ cm across all objects, highlighting the importance of finer voxel resolutions in self-calibration.

Different end-effector's point cloud sizes (i.e., $L$ in Sec. \ref{action:action}), including 60, 100, 600, 1000, and 2000, are tested for end-effector representation with a fixed SDF Resolution of 0.2 cm.
Objects with diverse contact characteristics show varying impacts on calibration accuracy, as demonstrated in Fig. \ref{graph:sim_test}(b).
For example, objects like the winerack and the shelf, which often make point contacts, are more sensitive to changes in the size of the point cloud.
In contrast, objects with more flat and larger surfaces, like tables and nightstands, demonstrate similar accuracies when different $L$ values are set.
Increasing the point cloud size generally leads to not only increased calibration accuracy but also a rise in average computational time for particle evaluation. As reported in Tab. ~\ref{tab:point_cloud}, the average computational time increases from $2.7s$ to $12s$ as $L$ increases from $60$ to $2000$. 
This indicates a noticeable trade-off between computation time and calibration accuracy.

            

\setlength\tabcolsep{4pt}
\begin{table}[h]
    \vspace{-0.3cm}
    \begin{center}
        \begin{tabular}{ |c|c|c|c|c|c| } 
            \hline
            $L$ & $60$ & $100$ & $600$ & $1000$ & $2000$\\
            \hline
            Time(s) & $2.7 \pm 0.3 $& $2.9 \pm 0.1$ &  $5.5 \pm 0.2$ & $7.8 \pm 0.8$ & $12.4 \pm 0.9$\\
            \hline
            \end{tabular}
            
    \end{center}
    \caption{Average Computation Time (s) for particle evaluation with different sizes of the point cloud ($L$).\label{tab:point_cloud}}
    \vspace{-0.3cm}
\end{table}

\subsubsection{Particle Evaluation Analysis}
\label{experiment:sigma}
This experiment focuses on particle weight assignment, governed by variance $\sigma_P$ detailed in Sec. \ref{geometric:eval}. 
Different values of $\sigma_P$ (0.001, 0.002, 0.005, 0.01, and 0.02) are investigated, as reported in Fig. \ref{graph:sim_test}(c).
A smaller $\sigma_P$ imposes a stricter criterion for particle rejection and makes the filter more similar to using binary contact representation, leading to suboptimal calibration and particle starvation. 
On the other hand, a larger $\sigma_P$ relaxes rejection criteria, keeping particles with a more erroneous hypothesis state and decreasing calibration accuracy. We observe a U-shaped graph with the lowest average final translational error of 0.53 cm at $\sigma_P = 0.005$.

\subsection{Real Robot Experiments}
\label{experiment:real}
In real-world deployments, calibration accuracy may vary both due to errors in the environment CAD models and measurement inaccuracies. 
Here, we validate the feasibility of the simulation-tested parameters in the real world. 

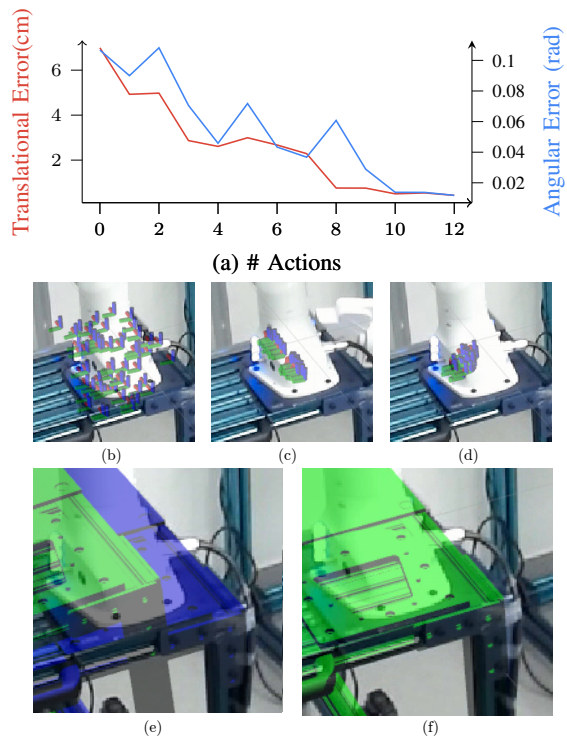
\begin{figure}[t]
    \centerline{
\begin{tikzpicture}

\definecolor{chocolate2196855}{RGB}{219,68,55}
\definecolor{darkgray176}{RGB}{176,176,176}
\definecolor{royalblue66133244}{RGB}{66,133,244}

\begin{axis}[
width=0.6\columnwidth,
height=0.25\columnwidth,
scale only axis,
axis x line*=bottom,
axis y line*=left,
tick align=outside,
tick pos=left,
x grid style={darkgray176},
xlabel={(a) \# Actions},
xmin=-0.6, xmax=12.6,
xtick style={color=black},
y grid style={darkgray176},
ylabel=\textcolor{chocolate2196855}{Translational Error(cm)},
ymin=0.105065359680105, ymax=7.32073296710728,
ytick style={color=black},
label style={font=\small}
]
\addplot [semithick, chocolate2196855]
table {%
0 6.99274807586059
1 4.92637485814085
2 4.9777828367944
3 2.86767089912244
4 2.610097647761
5 2.99434561450471
6 2.67268475742568
7 2.28775245541327
8 0.757866577590961
9 0.755036411126709
10 0.50028447929839
11 0.540347168210624
12 0.433050250926794
};
\end{axis}

\begin{axis}[
width=0.6\columnwidth,
height=0.25\columnwidth,
scale only axis,
axis x line*=bottom,
axis y line=right,
tick align=outside,
x grid style={darkgray176},
xlabel={(a) \# Actions},
xmin=-0.6, xmax=12.6,
xtick pos=left,
xtick style={color=black},
y grid style={darkgray176},
ylabel=\textcolor{royalblue66133244}{Angular Error (rad)},
ymin=0.00697916512902138, ymax=0.113021698403911,
ytick pos=right,
ytick style={color=black},
ytick={0.02,0.04,0.06,0.08,0.1},
yticklabels={0.02,0.04,0.06,0.08,0.1},
yticklabel style={anchor=west},
label style={font=\small}
]
\addplot [semithick, royalblue66133244]
table {%
0 0.106729902394602
1 0.090011782441195
2 0.108201583255052
3 0.0705447413679973
4 0.045832398455763
5 0.0717866378498152
6 0.0433672537427178
7 0.0367565438884304
8 0.0607501979197695
9 0.0289100817290676
10 0.0137850803929345
11 0.0137241649602453
12 0.01179928027788
};
\end{axis}

\end{tikzpicture}}
    \centerline{\includegraphics[width=0.8\columnwidth,keepaspectratio]{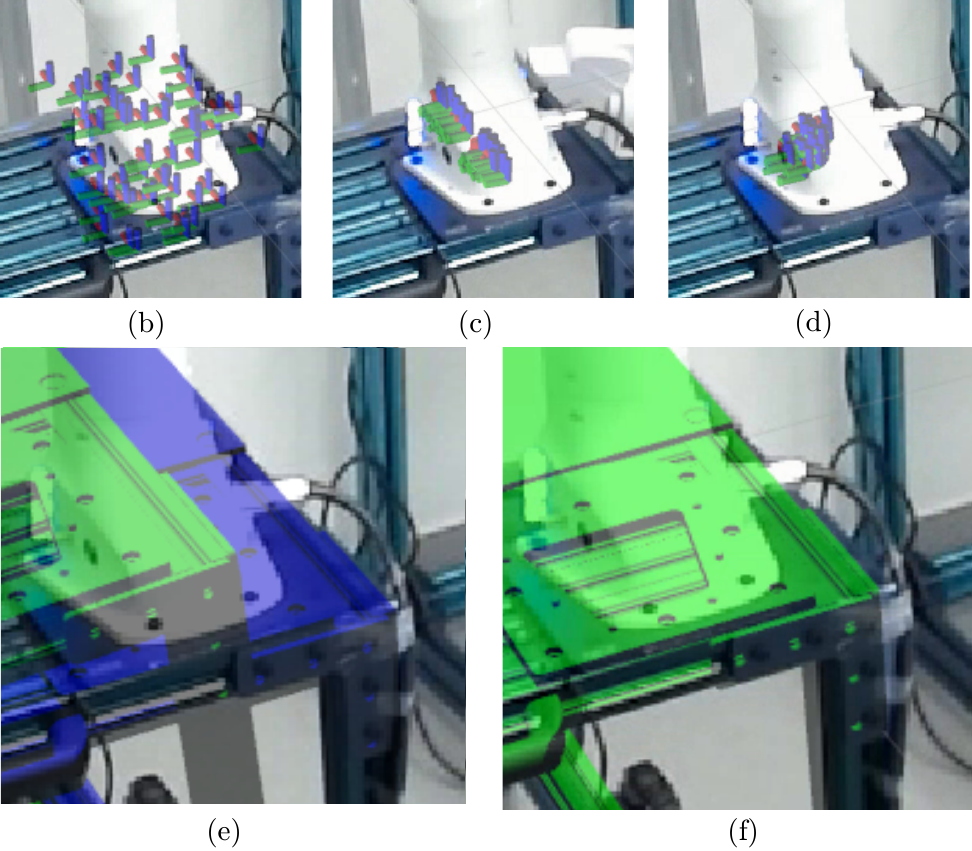}}
    \caption{(a) Translational (red) and rotational (blue) errors during the self-calibration. (b) initial hypothesis set  $\mathcal{X}_1$, (c) hypothesis set $\mathcal{X}_\tau$ updated based on the observation from touching and sliding action, (d) final hypothesis set $\mathcal{X}_t$. The green models overlay the initial guess (e) and the final estimation (f).}
    \label{graph:real_test}
    \vspace{-0.3cm}
\end{figure}

We use the optimal parameters obtained from simulation experiments, where $\Delta_{sdf}$ = 0.2 cm, $L$ = 600, and $\sigma_P$ = 0.005. 
The task involves self-calibrating the robot w.r.t. a table and a shelf in simulation and the real world, as depicted in Fig. \ref{pic:objects}. 
Each experiment is repeated 10 times.
An example run is presented in Fig. \ref{graph:real_test}, where we show the translational and rotational errors compared to the ground truth, the progression of the robot base's particle set $\mathcal{X}_t$, and the estimation overlays during the calibration process. 
The complete results are summarized in Tab. \ref{tab:real_result}.

The quality of actions, particularly the average number of contacts $J_t$ across all actions, significantly influences calibration accuracy. 
Calibrating a real-world shelf, disturbed by physically noisy touches against bolts, ridges, and narrow surfaces, results in fewer contacts per action (just under $4$) compared to simulation (above $5$). 
As a result, the final translational error in the real-world experiments is 0.72 cm, compared to 0.53 cm in the simulation. 
In contrast, calibrating a table without these variations in both simulation and the real world yields similar contact numbers, leading to comparable final translation and rotational errors.

We further see that self-calibration demonstrates adaptability to different geometric situations.
Calibrating the robot with respect to the table requires fewer actions for the particles to gain confidence and stability due to its relatively simple geometry.
The required number of actions is reduced from $30$ in the real world to $20$ in simulation, while the translational error remains around 0.5 cm.
Conversely, when calibrating a shelf, the translational error decreases from $0.7$cm (real-world) to $0.53$cm (simulation), while the average number of actions remains at 30.

Furthermore, we conduct the evaluation in a real-world application by applying the self-calibration framework w.r.t. a Baxter robot, as shown in Fig. \ref{pic:objects}. 
Given the absence of ground-truth information, a visual assessment is conducted. 
The results demonstrate a high self-calibration accuracy, underscoring the potential of the proposed framework in real-world robotic applications.
\setlength\tabcolsep{4pt}
\begin{table}[th]
    \vspace{-0.3cm}
    \begin{center}
        \begin{tabular}{ |c|c|c|c|c| } 
            \hline
            \multirow{2}{4em}{} &\multirow{2}{5em}{\centering Translational Error(cm)}  & \multirow{2}{6em}{\centering Rotational Error(${\scriptstyle 10 ^{-2}}$rad)}  & \multirow{2}{4em}{\centering \# Actions} & \multirow{2}{4em}{\centering \# Contact}\\
             &&&&\\
            \hline
            Table(Sim)& $0.51 \pm 0.17$ & $0.75 \pm 0.69$ & $21.8 \pm 5.8$ & $6.0 \pm 3.0$\\ 
            Table(Real)& $0.53 \pm 0.14$ & $ 1.10 \pm 0.41$& $17.2\pm 6.5$ & $ 6.1 \pm 3.0$\\ 
            Shelf(Sim)& $0.53 \pm 0.23$ & $0.50 \pm 0.20$ & $30.5 \pm 8.3$& $5.3 \pm 2.9 $\\ 
            Shelf(Real)& $0.72 \pm 0.30$ & $1.49 \pm 0.70$& $29.8 \pm 9.4$ & $ 3.9 \pm 1.0$\\ 
            \hline

            \end{tabular}

    \end{center}
    \caption{Experimental results for self-calibration using Table and Shelf from Fig. \ref{pic:objects} both in simulation and real environment.\label{tab:real_result}}
    \vspace{-0.3cm}
\end{table}

\section{Conclusion}
\label{conclusion}
This work proposes an exploratory action-based probabilistic estimation framework for self-calibrating the spatial relationship between the robot and its environment. 
Our approach gathers contact information and iteratively updates particle beliefs using exploratory touching and sliding actions. 
These actions are strategically chosen to enhance observation informativeness. 
An automatic verification mechanism is developed to ensure estimation quality and allows prompt termination when appropriate.
With extensive simulation and real-world experiments, we demonstrate the accuracy and effectiveness of our self-calibration framework in multiple different environments. 
Our particle update mechanism effectively converges hypothesized distribution towards the ground truth and consistently provides accurate calibration. 
In future work, we will extend our approach to self-calibrating more sensors, e.g., cameras, purely using the robot's exploratory actions without any human intervention or extra sensors.


\bibliographystyle{IEEEtran}
\bibliography{reference}

\end{document}